\documentclass{article}

\usepackage[final]{aaj2026}
\usepackage[numbers]{natbib}
\usepackage{graphicx} 
\usepackage{multirow}
\usepackage{subfig}
\usepackage{amsmath}

\usepackage[utf8]{inputenc} 
\usepackage[T1]{fontenc}    
\usepackage{hyperref}       
\usepackage{url}            
\usepackage{booktabs}       
\usepackage{amsfonts}       
\usepackage{nicefrac}       
\usepackage{microtype}      
\usepackage{xcolor}         

\title{MeTok: An Efficient Meteorological Tokenization with Hyper-Aligned Group Learning for Precipitation Nowcasting}

\author{%
  Qizhao Jin \\
  School of Artificial Intelligence\\ 
  University of Chinese Academy of Sciences\\
  \texttt{qizhao.jin@nlpr.ia.ac.cn} \\
  \
\And
 Xianhuang Xu \\
 China Meteorological Administration \\
 National Meteorological Centre \\
  \texttt{xunxianhuang@cma.gov.cn} \\
\And
 Yong Cao \\
 China Meteorological Administration \\
 National Meteorological Centre \\
  \texttt{caoyong@cma.gov.cn} \\
\And
Shiming Xiang \\
State Key Laboratory of Multimodal Artificial Intelligence Systems \\
Institute of Automation\\
  \texttt{smxiang@nlpr.ia.ac.cn} \\
\And
 Xinyu Xiao\\
   School of Artificial Intelligence\\ 
  University of Chinese Academy of Sciences\\
  \texttt{xinyu.xiao@nlpr.ia.ac.cn} \\
}

\begin{document}

\maketitle

\begin{abstract}
  Recently, Transformer-based architectures have advanced meteorological prediction. 
However, this position-centric tokenizer conflicts with the core principle of meteorological systems, where the weather phenomena undoubtedly involve synergistic interactions among multiple elements while positional information constitutes merely a component of the boundary conditions.
This paper focuses primarily on the task of precipitation nowcasting and develops an efficient distribution-centric \textbf{Me}teorological \textbf{Tok}enization (\textbf{MeTok}) scheme, which spatially sequences to group similar meteorological features. Based on the rearrangement, realigned group learning enhances robustness across precipitation patterns, especially extreme ones.
Specifically, we introduce the \textbf{Hy}per-\textbf{A}ligned \textbf{G}rouping \textbf{Transformer} (\textbf{HyAGTransformer}) with two key improvements: 1) The Grouping Attention (GA) mechanism uses MeTok to enable self-aligned learning of features from different precipitation patterns; 2) The Neighborhood Feed-Forward Network (N-FFN) integrates adjacent group features, aggregating contextual information to boost patch embedding discriminability.
Experiments on the ERA5 dataset for 6-hour forecasts show our method improves the IoU metric by at least 8.2\% in extreme precipitation prediction compared to other methods. Additionally, it gains performance with more training data and increased parameters, demonstrating scalability, stability, and superiority over traditional methods.
\end{abstract}

\section{Introduction}
Precipitation nowcasting is a short-range forecast of rainfall up to a few hours ahead, which often plays a crucial role in public services such as agricultural production, transportation planning, and altering natural hazardous events. 
In practical applications, meeting the timeliness and accuracy of the precipitation nowcasting algorithm, especially extreme precipitation prediction, has always been a big challenge. 
In the past few decades, the paradigm of numerical weather prediction (NWP) \cite{bauer2015quiet,lynch2008origins} has dominated precipitation forecasting, which is constructed on physical atmospheric rules for predicting the weather with current meteorological elements. 
Still, due to the high complexities of the mathematical models, which involve solving complicated partial differential equations, NWP methods could be computationally expensive \cite{sun2014use}. 

As a promising alternative, deep learning methods have been employed in the precipitation nowcasting task in recent years. 
Mainstream methods~\cite{ShiCWYWW15,PrecipitationNowcasting17,gao2024prediff} formulated precipitation nowcasting as a video prediction task. For instance, Gao et al.~\cite{DBLP:conf/nips/0001S0Z00Y22} introduced an efficient space-time Transformer \cite{Attention} for predicting future precipitation up to 60 minutes; Ravuri et al.~\cite{ravuri2021skilful} introduced DGMR, which generates spatiotemporally consistent predictions. They have been proven effective for low-intensity precipitation. 
However, the precipitation follows an extremely imbalanced long-tailed distribution. The dominance of head classes in the feature space leads classical models to overfit the abundant data \cite{zhang2023deep}, failing to capture the global distribution of rainfall. 
It could seriously damage the reliability of the learned model~\cite{bi2023accurate}.

\begin{figure}[t]
    \centering
    \includegraphics[width=\linewidth]{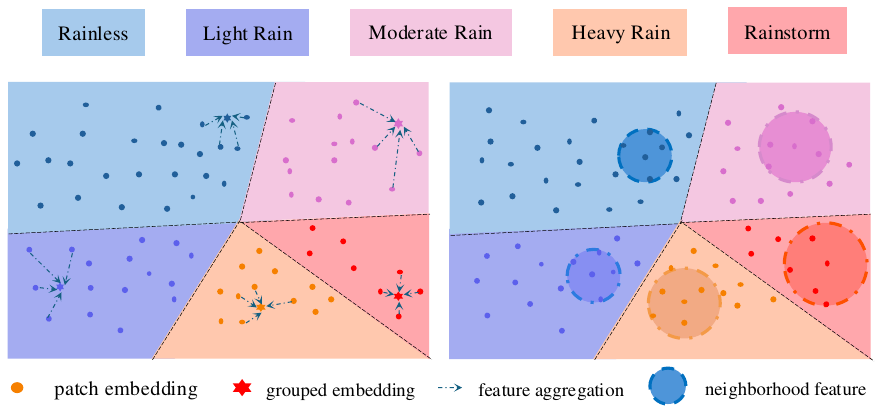}
    \caption{The weather phenomena essentially involve nonlinear interactions among multiple factors across multi-scale spatiotemporal domains. The MeTok scheme introduces the historical distribution of each embedding to gather information with similar patterns.}
    \label{fig:illustrate}
\end{figure}

The precipitation process is typically influenced by the spatial and temporal interactions of multi-scale atmospheric factors.
Inspired by the success of Transformer, this paper tries to utilize the attention mechanism to capture dependencies in both temporal and spatial dimensions~\cite{DBLP:conf/iclr/ParkK22}. 
Depending on the Transformers like ViTs \cite{DBLP:conf/iclr/DosovitskiyB0WZ21} or LLMs \cite{brown2020language,guo2025deepseek}, there is a tokenizer needs to be employed to sequentially convert visual or textual signals into a sequence of tokens. 
It indicates that the tokenizer plays a pivotal role that directly influences the performance of the Transformer models. 
Despite fully adopting approaches~\cite{DBLP:conf/nips/0001S0Z00Y22} based on visual or text tokenizations that have led to some success in precipitation nowcasting, their position-centric tokenization scheme overlooks that precipitation is the result of the interplay of various meteorological elements, with position information being just one of them \cite{harper2014weather}. 
The position-centric tokenization scheme combines meteorological features based on geographical position would merge different precipitation patterns and consequently impair the accuracy of precipitation nowcasting. 

This paper introduces a novel Meteorological Tokenization (\textbf{MeTok}) scheme that rearranges meteorological element patch embeddings based on historical precipitation distributions. As shown in Fig.~\ref{fig:illustrate}, this realignment follows a distribution-centric paradigm, grouping similar patch embeddings into non-overlapping clusters.
MeTok naturally captures dependencies between patch embeddings and group embeddings generated from intra-group patches. For long-tail precipitation modeling, its group learning encapsulates meteorological feature sets aligned with precipitation patterns, proving more efficient and robust than position-centric tokenizers, which let tail-sample patch embeddings be disrupted by global information.

Moreover, standard Transformers cannot support our distribution-centric MeTok. Thus, we propose the Hyper-aligned Grouping Transformer (HyAGTransformer) to integrate with MeTok. As Fig.~\ref{fig:illustrate} shows, it has two key traits: 1) enabling hyper-aligned discrimination of meteorological features across distinct precipitation patterns in patch embeddings; 2) leveraging similar precipitation pattern representations to explore aligned feature space information.
Structurally, HyAGTransformer includes a Grouping Attention (GA) mechanism and a Neighborhood Feed-Forward Network (N-FFN). GA captures dependencies between patch embeddings and their intra-group derived group embeddings, while reducing redundant features, particularly tail features, during attention computation. N-FFN uses each patch embedding’s neighborhood features to mitigate grouping-related information loss and explore contextual details, boosting discriminative power.

To accommodate MeTok and handle rearranged meteorological patch embeddings, we propose an encoder-translator-decoder framework using HyAGTransformer. 
Experimental results confirm the framework’s effectiveness: its long-tail precipitation nowcasting performance outperforms SOTA methods. On extreme precipitation ERA5 and WeatherBench datasets ~\cite{rasp2020weatherbench}, it achieves at least 4.7\% and 2.1\% higher IoU than baselines, respectively. 
Further tests on WeatherBench2 ~\cite{rasp2024weatherbench}demonstrate scalability—improving performance with more parameters and training data, and outperforming NWP methods.

\begin{figure}[t]
    \centering
    \subfloat[Patch embeddings in spatial order]
    {\includegraphics[width=0.45\columnwidth]{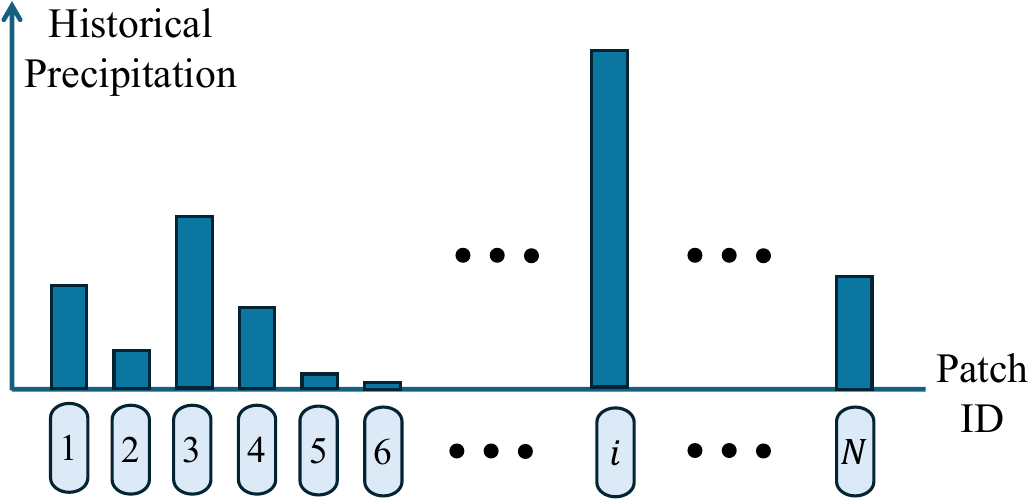}%
    \label{order}}
    \hfill
    \subfloat[Patch embeddings in precipitation order]
    {\raisebox{-.15\height}{\includegraphics[width=0.45\textwidth]{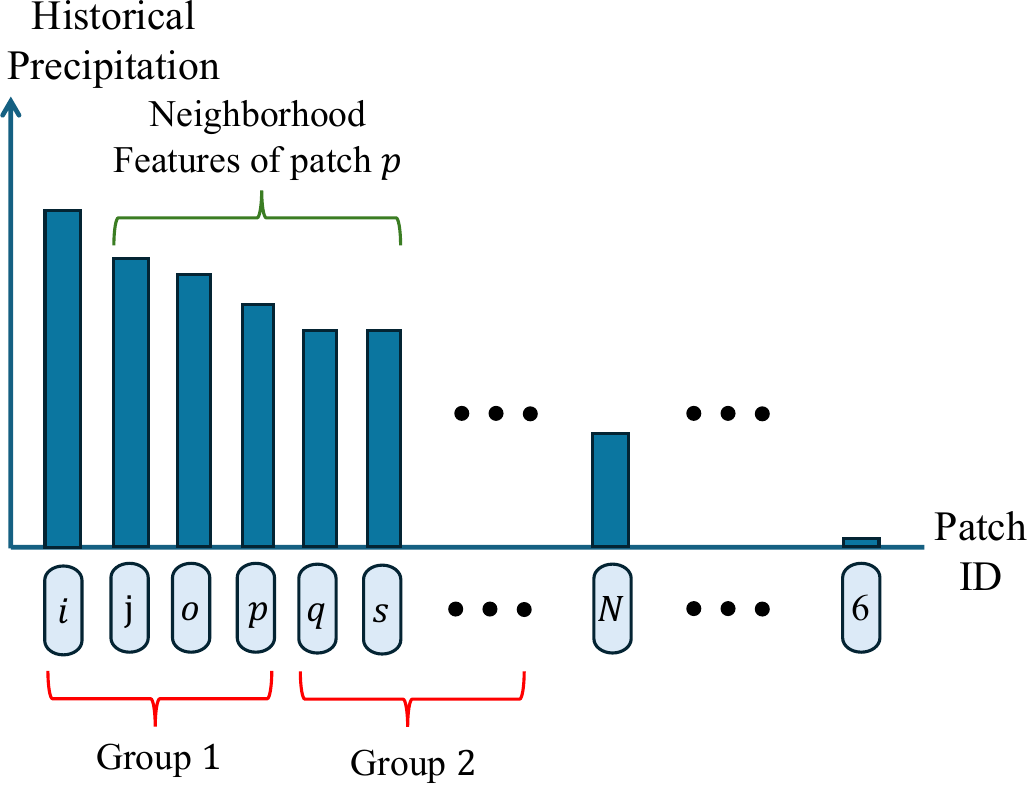}}%
    \label{weatherbench}}
    \hfill
    \caption{Patch embeddings are rearranged according to the historical precipitations.}
    \label{fig:order}
\end{figure}

\section{Related Work}

\subsection{Data Driven Based Precipitation Nowcasting}

Deep learning has enabled data-driven solutions for precipitation nowcasting, with mainstream radar-based methods framing it as spatiotemporal sequence prediction. ConvLSTM~\cite{ShiCWYWW15} pioneered this domain by replacing fully connected layers with convolutions to capture radar echo spatiotemporal features, spawning variants like TrajGRU \cite{PrecipitationNowcasting17}, MotionRNN \cite{DBLP:conf/cvpr/WuY0L21}, and PredRNN \cite{DBLP:conf/nips/WangLWGY17}. Subsequent works extended this paradigm using CNN-based \cite{DBLP:conf/cvpr/WuY0L21} or Transformer-based \cite{DBLP:conf/nips/0001S0Z00Y22} architectures.
Recent advances in generative models have also yielded radar-based approaches, such as DGMR \cite{ravuri2021skilful}, which simulates future radar samples conditioned on historical data.

\subsection{Tokenization}

Tokenization, a core preprocessing step in NLP and CV, converts raw data into structured formats. In NLP, it has advanced from basic word-based methods to sophisticated subword/character-level approaches. Early whitespace/punctuation-based segmentation struggled with morphological complexity and out-of-vocabulary terms.
Byte Pair Encoding (BPE) \cite{sennrich2015neural} was a key breakthrough, enabling effective handling of rare words. SentencePiece \cite{kudo2018sentencepiece} later generalized subword tokenization to raw text without language-specific preprocessing. More recently, Transformer models like BERT \cite{kenton2019bert} adopted WordPiece tokenization, showcasing its efficiency with diverse vocabularies.

In CV, tokenization gained traction with Vision Transformers (ViTs) \cite{DBLP:conf/iclr/DosovitskiyB0WZ21}, which split images into fixed-size patches as tokens. This lets ViTs use Transformer self-attention for global context modeling, though tokenization becomes computationally heavy for high-resolution images.
Swin Transformer \cite{liu2021Swin,liu2021swinv2} addressed this with hierarchical tokenization via shifted windows, cutting complexity while preserving performance. Discrete tokenization is another key approach, e.g., DALL·E \cite{ramesh2021zero} uses VQ-VAE \cite{van2017neural} to compress images into discrete tokens, bridging NLP and CV by aligning image representations with text tokens for unified multimodal modeling.

Overall, the tokenization in NLP and CV is position-centric, which effectively utilizes positional information. However, in the meteorological system, the evolution of weather phenomena involves nonlinear interactions among multiple factors, while positional information is merely one of the boundary conditions. Thus, this paper proposes the MeTok scheme tailored for meteorological forecasting tasks.

\section{Background}

\subsection{Problem Formulation}

This paper takes various meteorological elements as inputs to predict precipitation, including relative humidity, air temperature, and wind (the wind has two directions). These elements are represented by a continuous input $\mathbf{X}_t \in \mathbb{R}^{h \times w \times l}$ at time $t$, the spatial region is represented by an $h \times w$ grid, $l$ represents the number of variables. Given the length of $s$ previous observations, including the current one, the goal of precipitation nowcasting is to predict the most probable precipitation sequence in the future:
\begin{small}
\begin{equation}
    \label{E1}
    \hat{\mathbf{P}}_{t+1}, \hat{\mathbf{P}}_{t+2}, ..., \hat{\mathbf{P}}_{t+j}= F(\mathbf{X}_{t-s+1}, \mathbf{X}_{t-s+2}, ..., \mathbf{X}_{t}),
\end{equation}
\end{small}\noindent
where $\mathbf{P}_{t+1}\in\mathbb{N}^{h \times w}$ denotes precipitation between $t$-th observation and ${t+1}$-th observation in a certain area, $j$ is the length of the future, $s$ is the length of the history.

\subsection{Meteorological Tokenization}

In visual semantics, position information is critical for composing and interpreting visual content—hence positional or sequential patch embeddings in visual tokenization primarily convey spatial/sequential element positions.
Weather phenomena, by contrast, arise from nonlinear multi-factor interactions across multi-scale spatiotemporal domains, governed by physical laws \cite{frank2024characterizing,nguyen2023climax}. Position information is merely part of the boundary conditions; weather evolution depends on the spatiotemporal coupling of multidimensional elements. 

A detailed analysis of precipitation mechanisms, such as the Wegener-Bergeron-Findeisen and collision processes \cite{harper2014weather}, reveals that the formation of precipitation varies significantly under different meteorological conditions. Analogous weather patterns may occur at different locations while sharing similar dynamic-thermodynamic conditions.
Group-wise learning of meteorological features from historical data is methodologically sound with broad meteorological applications \cite{daly2008physiographically, hamill2015analog, lerch2017similarity}.
Thus, we propose an efficient meteorological tokenizer (MeTok) and identify precipitation nowcasting bottlenecks (explored further below). MeTok abandons traditional vision-based position-encoded nearby patching; instead, it rearranges meteorological map spatial sequences by distributional similarity/proximity, grouping similar features. To model this rearrangement, we introduce a historical distribution $\mathbf{H}$ to guide precipitation nowcasting, yielding Equation \ref{E1}:

\begin{small}
\begin{equation}
    \label{E1_1}
    \hat{\mathbf{P}}_{t+1}, \hat{\mathbf{P}}_{t+2}, ..., \hat{\mathbf{P}}_{t+j}= F({\mathbf{X}}_{t-s+1}, {\mathbf{X}}_{t-s+2}, ..., {\mathbf{X}}_{t}, \mathbf{H}),
\end{equation}
\end{small}\noindent
where $\mathbf{H}$ represents the historical distribution of meteorological features, which can provide a stable and consistent reference for clustering similar meteorological features.

\section{Our Approach}
\subsection{Preliminary}
For the MeTok, this paper introduce a encoder-translator-decoder framework. The full pipeline $F()$ can be split to an encoder $\mathcal{E}$, a translator $\mathcal{T}$ and a decoder $\mathcal{D}$:
\begin{equation}
    [\mathbf{O}_*] = \mathcal{E}([\mathbf{X}_*],\mathbf{H}),~[\mathbf{V}_*] = \mathcal{T}([\mathbf{O}_*], \mathbf{H}),~[\mathbf{P}_*] = \mathcal{D}([\mathbf{V}_*]),
\end{equation}
where $[\mathbf{X}_*]$, $[\mathbf{O}_*]$, $[\mathbf{V}_*]$ and $[\mathbf{P}_*]$ refer to the input and output sequences of each module. 

\begin{figure*}[t!]
    \centering
    \includegraphics[width=\textwidth]{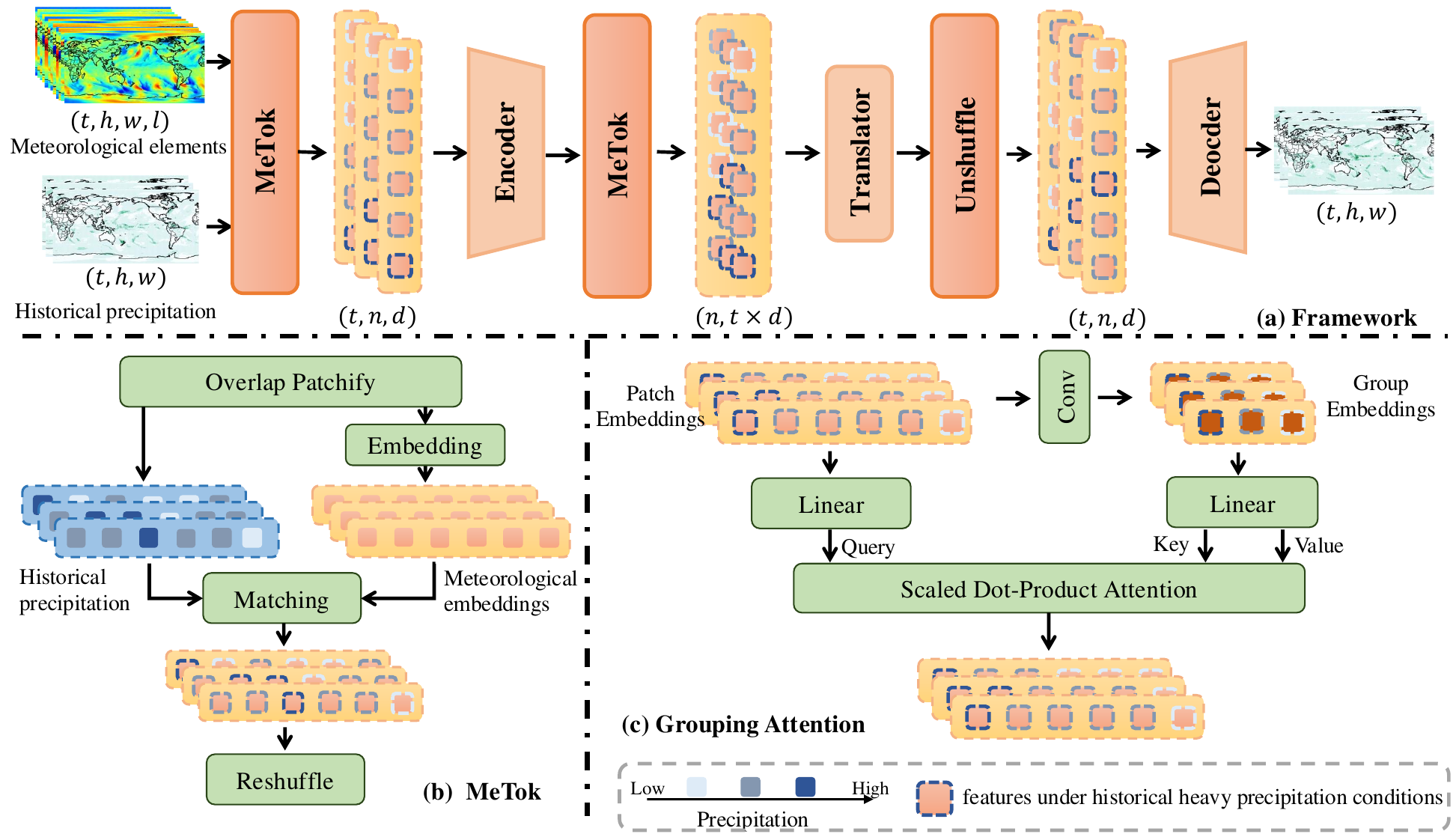}
    \caption{(a) The encoder-translator-decoder framework. The encoder models the spatial features with the rearranged patch features. The translator learns spatiotemporal dynamics. The decoder integrates the spatiotemporal information and local features. (b) The meteorological embeddings are rearranged based on the historical precipitations in the MeTok scheme. (c) Within the HyAGTransformer, the GA mechanism captures dependencies between embeddings and grouped features.}
    \label{fig:detailed}
\end{figure*}

\subsection{Hyper-Aligned Grouping Transformer}

{\bfseries Patchify.} To input the meteorological elements into Transformer, we follow the ViT framework and transfer them into spatial patches. The modeling process is defined as: $\mathbf{Z}_t^p = \mathcal{P}(\mathbf{X}_t)$, where $\mathcal{P}$ denotes the linear patchify operation which consists of a linear projection with the added positional embeddings. $\mathbf{Z}_t^p \in \mathbb{R}^{\frac{h}{p} \times \frac{w}{p} \times d}$ is the patch embedding which projects into $d$ dimensions, $(p, p)$ is the resolution of each patch. For clarity, this paper set $n = \frac{h}{p} \times \frac{w}{p}$.

{\bfseries Spatiotemporal Positional Embedding.} 
To encode the patches with spatiotemporal information, we adopt solar elevation angles $\alpha_t$ via the spatiotemporal positional embedding layers. The solar elevation angle is the angle between sunlight and the horizontal plane, which offers key benefits: it directly relates to longitude, latitude, and time, calculated as follows:
\begin{small}
\begin{equation}
    \label{E4}
        \alpha_t = \arccos(\sin \Phi \sin \delta + \cos \Phi \cos \delta \cos s_t),
\end{equation}
\end{small}\noindent
where $s_t$ is the local solar time, $\Phi$ is the local latitude, and $\delta$ is the current declination of the sun; additionally, solar radiation is the primary energy source driving atmospheric motion, and solar elevation angles reflect its intensity.

{\bfseries Group Embedding.} 
This paper denotes the group embeddings $\mathbf{E}_t \in \mathbb{R}^{m \times d}$, where $m$ is the size of the group embedding space. 
In MeTok, as shown in Fig.~\ref{fig:detailed} (b), the group embeddings are derived from patch embeddings with similar historical distribution.
Specifically, we set historical distribution as historical precipitations for simplicity, e.g., $\mathbf{H}=(\mathbf{P}_{t-s+1}, \mathbf{P}_{t-s+2}, ..., \mathbf{P}_{t})$.
To facilitate implementation, patch embeddings are rearranged in descending order along the axis of historical precipitations as follows:
\begin{equation}
\label{E5}
\mathbf{\hat{Z}}^p_t = {\rm Rearrange}(\mathbf{Z}^p_t \mid \mathbf{H}),
\end{equation}
where $\mathbf{\hat{Z}}^p_t \in \mathbb{R}^{n \times d}$ is the rearranged patch embeddings based on the historical precipitations $\mathbf{H}$. 
The rearranged patch embeddings $\mathbf{\hat{Z}}^p_t$ are then divided into static, non-overlapping groups, with embeddings within each group aggregated into group embedding.
The grouping and aggregating intra-group information operations are implemented with 1D convolutions, wherein the stride and kernel size are set to $n/m$, which is the same as the number of patches per group. Given the rearranged patch embedding $\mathbf{\hat{Z}}^p_t$, the group is generated as follows:
\begin{equation}
\mathbf{E}_t = {\rm Conv1d}(\mathbf{{\hat{Z}}}^p_t, {\rm kernel \ size}=n/m, {\rm stride}=n/m).
\end{equation}

{\bfseries HyAGTransformer Structure.}
To deal with the rearranged grouping of meteorological features, we proposed the Hyper-Aligned Grouping Transformer (HyAGTransformer) to design a grouping attention (GA) mechanism and neighborhood feed-forward network (N-FFN). The GA mechanism captures the dependence between the patch and group embeddings to replace traditional self-attention. 
Different from plain self-attention, the GA mechanism generates key and value from the group embeddings $\mathbf{E}_t$ while mapping the query from the patch features $\mathbf{\hat{Z}}^p_t$ as follows:
\begin{align}
\mathbf{Q} &= \mathbf{W}_q \cdot \mathbf{\hat{Z}}^p_t, \\
\mathbf{K} &= \mathbf{W}_k \cdot \mathbf{E}_t, \\
\mathbf{V} &= \mathbf{W}_v \cdot \mathbf{E}_t, 
\end{align}
where the $\mathbf{W}_q$, $\mathbf{W}_k$, and $\mathbf{W}_v$ are parameter matrices for computing query, key, and value vectors, respectively. 

GA mechanism possesses distinct advantages: First, GA mechanism effectively reduces the computational complexity from $\mathcal{O}(n^2)$ to $\mathcal{O}(nm)$ where $m < n$. 
Furthermore, the GA mechanism facilitates hyper-aligned discrimination of meteorological features across distinct precipitation patterns in the embedding space.

Although group embedding effectively aggregated features within aligned precipitation patterns, it inevitably induced cross-group misalignment losses during the aggregation process.
Moreover, the grouping strategy adopted in the GA mechanism will discretize the continuous feature space. 
This is unfavorable for encoding continuous meteorological features.
Hence, HyAGTransformer utilizes the neighborhood features of each embedding to explore contextual information in the feature space.
As illustrated in Fig.~\ref{fig:order}, the neighborhood features of embedding $p$ are defined as the $n/m-1$ nearest features along the rearranged embedding. 
We design the N-FFN, which incorporates 1D convolutions within the feed-forward network to capture the local dependencies among neighborhood features, which can be formulated as follows:
\begin{equation}
\mathbf{Z}^p_t = {\rm MLP}({\rm GELU}({\rm Conv1d}({\rm MLP}(\mathbf{Z}^p_t)))).
\end{equation}
In contrast to plain FFN, N-FFN delivers aligned contextual awareness within the feature space through minimal parametric overhead, while mitigating GA-induced cross-group misalignment from the grouping strategy.

\begin{table}[!tbp]
\centering
{
    \begin{tabular}{l|c|c|c|c}
        \toprule
        Model & Params(M) & MACs(G) & TS~$\uparrow$ & IoU~$\uparrow$ \\
        \midrule
        ConvLSTM  & 14.79 & 463.74 & 20.68 & 37.83 \\
        ConvGRU  & 17.19 & 604.8 & 20.41 & 38.03  \\
        TrajGRU  & {12.72} & 720.64 & 19.50 & 36.09  \\
        PredRNN  & 25.84 & 1103.64 & 17.95 & 36.58  \\
        PFST  & 40.65 & 472.97 & 18.95 & 38.05 \\
        SCCN  & 48.1 & 2990.48 & 19.77 & 38.01  \\
        SimVP  & 14.08 & 419.44 & 20.52 & 39.84   \\
        SimVP+gSTA & 10.48 & 330.96 & 20.27 & 39.46  \\
        TAU & 10.05 & 318.78 & 20.58 & 39.76 \\ 
        Ours & 1.42 & {387.06} & \textbf{23.33} & \textbf{44.41} \\
        \bottomrule
    \end{tabular}
} 
\caption{Overall performances of our framework on ERA5 dataset. The results are the mean values covering all categories, where the values in bold are the best.}
\label{tab:overallera5}
\end{table}

\subsection{Framework}

Depending on MeTok, the HyAGTransformer is applied to construct an encoder-translator-decoder framework. 



{\bfseries Encoder.} Given the input tensor $\mathbf{X}$ with size $(t, h, w, l)$, we divide it into overlapped patches~\cite{DBLP:conf/iccv/WangX0FSLL0021}. Patch embeddings $\mathbf{Z}^p=[\mathbf{Z}^p_{t-s+1}, \mathbf{Z}^p_{t-s+2}, ..., \mathbf{Z}^p_{t}]$ are encoded by the Patchify module with size $(t, n, d)$. 
Then, this paper employs $N_E$ HyAGTransformer blocks to capture spatial dependency of patch embeddings $\mathbf{Z}^p$ following the MeTok scheme. 
For the $N_e$-th HyAGTransformer block, patch embeddings $\mathbf{Z}^p$ are rearranged in descending order along the axis of corresponding historical precipitations as follows:
\begin{equation}
    \mathbf{\hat{Z}}^p_i = {\rm Rearrange}(\mathbf{Z}^p_i \mid \mathbf{P}_{i}), \qquad i \in (t-s+1, t)
\end{equation}
Due to the varying historical precipitations $\mathbf{P}_{i}$ along the temporal axis, the groupings at each moment are not uniform.
The group embeddings $\mathbf{E}$ aggregate features under similar historical precipitations by partitioning the rearranged patch embeddings $\mathbf{\hat{Z}}^p$ into groups of equal quantities. 
Within the HyAGTransformer blocks, the GA layer is utilized to model the dependency between the patch embedding $\mathbf{\hat{Z}}^p$ and group embedding $\mathbf{E}$ and the N-FFN performs feature interaction with neighborhood features as follows: 
\begin{align}
    \mathbf{\hat{Z}}^p &= \mathbf{\hat{Z}}^p + {\rm LayerNorm}({\rm GA}(\mathbf{\hat{Z}}^p, \mathbf{E})) \\
    \mathbf{\hat{Z}}^p &= \mathbf{\hat{Z}}^p + {\rm LayerNorm}(\text{N-FFN}(\mathbf{\hat{Z}}^p)) 
\end{align}
On the top of the HyAGTransformer block, rearranged patch embeddings $\mathbf{\hat{Z}}^p$ are unshuffled to facilitate subsequent processing. 
\begin{equation}
    \mathbf{{Z}}^p_i = {\rm Unshuffle}(\mathbf{\hat{Z}}^p_i \mid \mathbf{P}_{i}), \qquad i \in (t-s+1, t)
\end{equation}

{\bfseries Translator.} 
In the translator, we employ $N_T$ HyAGTransformers for modeling temporal evolution. 
To capture the temporal global variations, we resize the output features of the encoder $\mathbf{O}\in \mathbb{R}^{t, n,d}$ into $(n, t\times d)$. 
This structure offers two benefits: it boosts computational efficiency by reducing the burden of encoding global spatiotemporal relationships simultaneously, and it establishes hyper-aligned temporal foundations for spatial dependencies—these require global temporal feature integration, not isolation to the present moment, when capturing inter-patch dependencies.

The rearranged data structure requires that the Metok mechanism also be changed accordingly. 
Among the $N_t$-th HyAGTransformer block, patch embeddings $\mathbf{Z}^p\in \mathbb{R}^{n, t\times d}$ are rearranged with a temporally shared grouping strategy as follows:
\begin{equation}
    \mathbf{\hat{Z}}^p_{t-s+1}, ..., \mathbf{\hat{Z}}^p_t = {\rm Rearrange}(\mathbf{Z}^p_{t-s+1},..., \mathbf{Z}^p_t,\mid \mathbf{P}_{t}).
\end{equation}

\begin{table}[!t]
\centering
{
    \begin{tabular}{l|c|c|c|c}
        \toprule
        Model & Params(M) & MACs(G) & TS~$\uparrow$ & IoU~$\uparrow$  \\
        \midrule
        ConvLSTM & 14.03 & 309.16 & 14.86 & 28.81   \\
        ConvGRU  & 17.19 & 403.2 & 14.34 & 26.99    \\
        TrajGRU  & {12.72}  & 480.43 & 14.83 & 28.69  \\
        PredRNN  & 25.84 & 735.76 & 14.52 & 28.64  \\
        PFST  & 40.65 & 315.31 & 11.68 & 26.32 \\
        SCCN  & 48.1 & 1993.71 & 14.82 & 29.64  \\
        SimVP & 14.08 & 279.63 & 17.93 & 35.47   \\
        SimVP+gSTA& 10.48 & 220.64 & 17.71 & 35.19  \\
        TAU & 10.05 & 212.52 & 17.31 & 34.95 \\
        Ours & 1.42 & {258.04} & \textbf{19.33} & \textbf{37.56} \\
        \bottomrule
    \end{tabular}
} 
\caption{Overall performances of our framework on the WeatherBench dataset. The results are the mean values covering all categories, where the values in bold are the best.}
\label{tab:overallweatherbench}
\end{table}

{\bfseries Decoder.} The decoder combines translator-derived spatiotemporal features with local spatial features to predict upcoming precipitation. It first restores spatiotemporal features’ spatial positions, aligning patches with their pre-grouping arrangement.
Utilizing transposed convolutional blocks, we perform upsampling of the spatiotemporal features, generating feature maps $\mathbf{{Z}}_{g}$ that match the original dimensions.
Another convolutional block integrates $\mathbf{{Z}}_{g}$ and local features $\mathbf{Z}_{l}$ for the final prediction.

\section{Experiments}
\subsection{Dataset}

\begin{table}[t]
\centering
\setlength{\tabcolsep}{0.5mm}
\begin{tabular}{l|c|c|c}
\toprule
                               & ERA5 & WB & WB 2 \\ 
\midrule
Spatial resolution             & $0.25^\circ$     &  $1.40625^\circ$  &    $1.5^\circ$                        \\
Temporal resolution            &   1 hr   &  1 hr  & 6hr                          \\
Vairables                &  12    & 12   &    79                      \\
\multicolumn{1}{l|}{\multirow{2}{*}{Spatial range}} & $0^\circ$ $\sim$ $55^\circ$N & \multirow{2}{*}{global} & \multirow{2}{*}{global}         \\
\multicolumn{1}{l|}{}                               & $140^\circ$E $\sim$ $70^\circ$W &                   &                           \\
Training data                  & 2016$\sim$2018 & 2014$\sim$2016 & 1989$\sim$2019                           \\
Validation data                & 2019 & 2017 & 2020                           \\
Test data                      & 2020 & 2018 & 2021                           \\ 
\bottomrule
\end{tabular}
\caption{The experiment setting of ERA5, WeatherBench, and WeatherBench 2.}
\label{fig:settings}
\end{table}

This section mainly introduces the detailed settings for the datasets used in this paper, including  regional ERA5, WeatherBench, and WeatherBench2, as shown in Table~\ref{fig:settings}. 

For the ERA5 dataset, we take 12 hourly variables as input to predict precipitation, including relative humidity, air temperature, u-component of wind, and v-component of wind at 500hPa, 850hPa, and 1000hPa levels. The spatial resolution of the original data is $0.25^\circ$ covering $1440\times721$ grid cells. Due to the large amount of computation required when directly using the raw data, we selected a subset for our research. The study area encompasses a $221 \times 281$ grid, spanning from 140 degrees east to 70 degrees west and 55 degrees north to the equator. Along the temporal dimension, we use the data from 2016 to 2018 as the training set, the data from 2019 as the validation set, and the data from 2020 as the test set.

The WeatherBench dataset serves as a benchmark for global meteorological data. This paper also utilizes 12 hourly variables as input. 
At the spatial axis, a horizontal resolution of $1.40625^\circ$ with a $128 \times 256$ grid is selected. 
The WB dataset covers five years from 2014 to 2018 at the temporal axis. 
The meteorological data from the first three years constitute the training set, while the data from the fourth and fifth years serve as the validation and testing datasets, respectively

To further explore the model's scalability and compare it with traditional methods, this paper uses the WeatherBench2 dataset which provides a set of baselines consisting of ECMWF’s operational IFS mode and data-driven model. For the WB2 dataset, this paper chooses 79 variables that span six hours, including relative humidity, air temperature, u-component of wind, v-component of wind, vertical velocity, geopotential in 13 levels, and cloud cover. At the training stage, we use 30 years of data from 1989 to 2019; during the validation and testing phases, this paper uses data from 2020 and 2021, respectively.

\subsection{Implementation Details}
On the ERA5 and WB datasets, we evaluate the model's performance for precipitation intensity nowcasting. This study predicts hourly precipitation intensity for the upcoming 6 hours based on historical meteorological elements spanning a length of $j$. The future prediction involves cumulative precipitation over two periods, such as the precipitation between $t$ and $t+1$. We employ the intersection over union (IoU) and the threat score (TS) to assess the proposed method comprehensively. The precipitation threshold is outlined in the Appendix Section, and the metric derivations mentioned earlier can be found in the Appendix Section.

To further verify the model's effectiveness, we compared the proposed method with the IFS model and SOTA data-driven methods on the WB2 dataset for short-term precipitation regression tasks. This research predicts the cumulative precipitation every 6 hours for the next 36 hours. The root mean squared error (RMSE) is used to evaluate the model's performance. The evaluation code and data are available in the WB2 dataset.

The patch features are extracted from the concatenation of multimodal data. We implement the overlapped patch embedding with a convolutional layer whose stride is 4, i.e. $p=4$. When conducting the grouping procedure, both within the encoder and decoder, the number of patches within each group is consistently configured to be 8. In this paper, the number of HyAGTransformer in the encoder is set to 4, and the number of layers of the translator is set to 2. The framework is trained from scratch.

The training process is performed by utilizing the AdamW optimizer \cite{DBLP:conf/iclr/LoshchilovH19} for 50 training epochs. We use a batch size of 32 for both datasets. The learning rate is set to an initial value of $10^{-3}$ and follows an exponential LR schedule with gamma 0.9. At each epoch, the validation set is used to evaluate the training model, and the best performance model is selected for testing. All experiments are implemented with Pytorch \cite{paszke2017automatic} on 4 Titan RTX 3090 with 24G memory.

\subsection{Comparision with Other Methods}

\begin{table}[t]
\centering
\setlength{\tabcolsep}{0.8mm}
{
    \begin{tabular}{lcccccc}
        \toprule
        \multirow{2.5}*{Method} & \multicolumn{6}{c}{Prediction time (h)}\\
        \cmidrule(lr){2-7}
        & 0$\sim$6& 6$\sim$12 & 12$\sim$18 & 18$\sim$24 & 24$\sim$30 & 30$\sim$36\\
        \midrule
         IFS HRES& 1.39& 1.28&1.27&1.45&1.42&1.57\\
         IFS ENS mean& 1.29&1.15 &1.11&1.30&1.23&1.39\\
         Ours& 0.84 &1.03& 1.22& 1.35& 1.49&1.58 \\
        \bottomrule
    \end{tabular}
} 
\caption{Performance of the proposed method on the WeatherBench 2 dataset.}

\label{tab:wb2_reg}
\end{table}

On the ERA5 and WB datasets, several other methods were compared to evaluate the effectiveness of the proposed framework. 
All comparison methods adopt the same input format as ours for a fair comparison. Some methods, like PFST, have been slightly adjusted to suit the datasets and predictions of our task.

As shown in Tab.~\ref{tab:overallera5} and Tab.~\ref{tab:overallweatherbench}, our framework achieved SOTA performance. On the ERA5 dataset, our framework demonstrates an improvement of at least 4.57\% on the IoU metric and 1.75\% on the TS metric. On the WB dataset, compared to other methods, our framework demonstrates an improvement of at least 2.37\% on the IoU metric and 1.4\% on the TS metric. 
Compared to others, it outperforms with only marginal extra computational overhead, showing MeTok’s performance gains aren’t due to additional parameters.

This paper then conducts experiments on the WB2 dataset to evaluate the proposed model on the IFS model and SOTA data-driven methods. 
As shown in Tab. \ref{tab:wb2_reg}, our MeTok outperforms the IFS model over the next 0-18 hours prediction. 
This further demonstrates the model's effectiveness in precipitation nowcasting.
Tab. \ref{tab:chap5_scal} exhibits that the proposed MeTok can improve performance by increasing the model parameters and expanding the training data scale. 
Therefore, we believe that the model's prediction quality will be further boosted by increasing computational resources.

\begin{table}[t]
\centering
\begin{tabular}{cccc}
\toprule
Training data & \# Var. & \# Param. & RMSE \\ 
\midrule
2009$\sim$2019 & 52 & 4.82M & 1.91 \\
1999$\sim$2019 & 52 & 8.15M & 1.48 \\
1989$\sim$2019 & 52 & 8.15M & 1.34 \\
1989$\sim$2019 & 79 & 8.15M & 1.25\\ 
\bottomrule
\end{tabular}
\caption{Study on the scalability of the proposed method on the WeatherBench 2 data set.}

\label{tab:chap5_scal}
\end{table}
\subsection{The long-tailed precipitation prediction}
\begin{table}[t]

	\centering	
	\setlength{\tabcolsep}{1mm}
	{
	\begin{tabular}{lccccc}
		
		\toprule
		Model & RL & LR & MR & HR & RS \\
		\midrule
		PFST  & 88.13 &  50.12 & 22.88 & 17.27 & 12.04 \\
		SCCN  &  87.13 &  51.42 & 20.16 & 17.5 & 13.85 \\
            SimVP & 87.28 & 50.97 & 25.89 & 21.02 & 14.04 \\
            SimVP+gSTA & 87.23 & 51.71 & 25.13 & 19.36 & 13.88 \\
            TAU & {88.41} & 52.30 & 26.33 & 19.31 & 12.58 \\
            Ours  & \textbf{88.52} & \textbf{54.25} & \textbf{30.04} & \textbf{25.01} & \textbf{22.25} \\
		\bottomrule
	\end{tabular} 
	}
\caption{Results on the ERA5 dataset of each precipitation nowcasting IoU for multiple levels. We report IoU (\%) for each level.}

\label{tab:multi}

\end{table}

\begin{table}[t]

	\centering	
	\setlength{\tabcolsep}{1mm}
	{
	\begin{tabular}{lccccc}
		
		\toprule
		Method & RL & LR & MR & HR & RS \\
		\midrule
	Dice Loss  & 87.08 & 50.31 & 25.89 & 20.20 & 14.79\\
        Seesaw Loss & 88.28 & 54.18 & 26.59 & 18.82 & 17.20 \\
        Recall Loss  & 88.12 & 53.93 & 29.22 & 20.92 & 19.82 \\
        \midrule
        Ours(Dice Loss)  & \textbf{88.52} & {54.25} & {30.04} & \textbf{25.01} & \textbf{22.25} \\
        Ours(Recall Loss)  & {88.39} & \textbf{56.89} & \textbf{31.30} & {24.72} & {20.73} \\
	\bottomrule
	\end{tabular} 
	}
     \caption{The comparison of IoU metric of different long tail learning methods on the ERA5 dataset.}

\label{tab:chap5_compare}

\end{table}
The comparison of multiple rain levels on the ERA5 dataset is presented in Tab. \ref{tab:multi}. In a comprehensive view, the proposed framework consistently demonstrates higher predictive accuracy than existing models across various precipitation intensities.
Regarding predictions for tail categories, the performance improvement is even more pronounced. In the moderate and heavy rain classes, our framework achieves a predictive accuracy improvement of at least $3.7\%$ and $4.0\%$, respectively, on the IoU metric. The forecasting accuracy in the heavy rainfall class demonstrates the most substantial improvement, with a remarkable enhancement of 8.2\%.
We argue that the cause behind the elevated predictive accuracy for tail classes lies in the amplified robustness of the tail features, owing to the information aggregation amongst patch features sharing akin precipitation patterns in MeTok.
Moreover, while the proposed method only marginally improves the prediction accuracy of head classes, it demonstrates that grouping tail sample features does not compromise the modeling ability of head classes.

This paper also compares the proposed method with several commonly used long-tail losses, including Dice Loss~\cite{DiceLoss}, Seesaw Loss~\cite{DBLP:conf/cvpr/WangZZCPGCLLL21}, and Recall Loss~\cite{DBLP:conf/icra/TianMSCK22}. 
These loss functions are employed to provide supervised information for training a benchmark model that replaces the HyAGTransformer with a plain Transformer.
As shown in Tab~\ref{tab:chap5_compare},
it is evident that, irrespective of the loss function used for supervision, the benchmark model consistently underperforms compared to the proposed method across all precipitation intensity levels. 
Although these loss functions improve the model's predictive ability, the performance gain brought by supervised information is relatively small for the tail classes such as "HR" and "RS" compared to the proposed method.
The main reason is that the precipitation follows an extremely unbalanced distribution, making it difficult to enhance the model's perception ability of tail classes by adjusting the weight of different category losses.
It is noteworthy that the rearranged group learning in MeTok is parallel to existing long-tail learning technologies.
As shown in Tab.~\ref{tab:chap5_compare}, combining the proposed method with other approaches further enhance the model's performance.

\begin{table}[!tbp]
\centering
{
    \begin{tabular}{ccccc}
        \toprule
        MeTok & GA & N-FFN & IoU~$\uparrow$ & TS~$\uparrow$ \\
        \midrule
         &  &  &40.93&21.04\\
        \checkmark &  &  &41.19&21.22\\
        \checkmark & \checkmark &  &42.74&21.51\\
        \checkmark &  & \checkmark &43.93&21.96\\
         & \checkmark & \checkmark &40.79&20.98\\
        \checkmark &\checkmark & \checkmark &\textbf{44.41}&\textbf{22.33}\\
        \bottomrule
    \end{tabular}
} 
\caption{Ablation studies on the effectiveness of each ingredient of HyAGTransformer on the ERA5 dataset.}

\label{tab:component}
\end{table}

\begin{figure}[t]
    \centering
    \includegraphics[width=250pt]{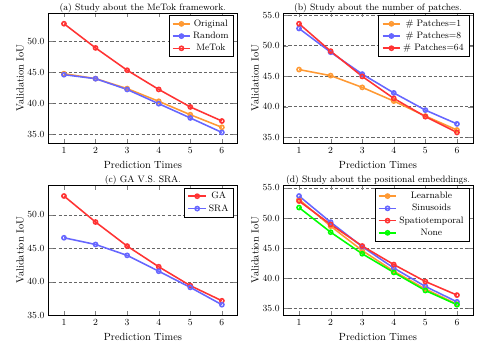}
    \caption{The ablation study of the proposed framework. We conducted a comprehensive analysis of the proposed framework, examining it from various facets.}
    \label{fig:comparison}

\end{figure}

\subsection{Ablation Study}

\noindent {\bf{Influence of the three ingredients on performance.}} We conduct experiments to comprehensively analyze the influence of the three ingredients on performance, which is illustrated in Tab. \ref{tab:component}. In contrast to the baseline model, which lacks three ingredients, our model gains a 3\% improvement on the mIoU metric. Suppose we were to rearrange the patch features solely based on historical precipitations without accounting for the feature aggregation within groups. In that case, the performance of this approach remains nearly akin to that of the baseline model. Following the rearrangement of patches, the option for either GA or N-FNN individually fails to yield optimal performance. In the absence of sorting, even with the incorporation of GA and N-FNN, no discernible enhancement in performance can be observed. Hence, we conclude that reordering patch features and feature aggregation are indispensable components.


\noindent {\bf{Study about the MeTok strategy.}} We investigate the influence of diverse ordering strategies on the model's performance. Three approaches are primarily examined: original, random, and ordering based on historical precipitation patterns. Fig. \ref{fig:comparison} (a) shows that arranging patch features according to precipitation yields a performance gain with an evident margin in contrast to the preceding two methodologies. 
We argue that the principal cause is that post-sorting, patches with analogous precipitation patterns become more amenable to aggregation, thereby generating more discriminative features under our MeTok scheme. 

\noindent {\bf{Study about the number of patches.}} Exploring the impact of the number of groups on model performance by varying the number of patches within each group. As shown in Fig. \ref{fig:comparison} (b), the model's performance exhibits a variation as the number of groups changes. Although the model does not excel at all periods, its overall performance is superior when the number of patches equals 8. A scarcity of patches within a group compromises the capacity to model robust intra-group features, while an excess of patches can introduce confusion among patches with distinct precipitation patterns.



\noindent {\bf{GA V.S. SRA.}} We have compared the GA with Spatial-Reduction Attention (SRA) \cite{DBLP:conf/iccv/WangX0FSLL0021}. The former performs feature aggregation along the rearranged feature space, while the latter extracts local features in the spatial axis. We conducted the experiments to facilitate a comprehensive comparison between these two. The inappropriateness of SRA for rearrangement patches stems from the deviation from their initial spatial arrangement. As shown in Fig. \ref{fig:comparison} (c), the SRA performed on the original patches is inferior to the GA.



\noindent {\bf{Study about the spatiotemporal positional embeddings.}} To evaluate the effectiveness of the spatiotemporal positional embeddings, an experiment is conducted by replacing them with learned positional embeddings, sinusoidal positional encodings or null embeddings. As shown in Fig. \ref{fig:comparison} (d), while the learned and sinusoidal positional encodings exhibit a slight advantage in the short term, the spatiotemporal positional embeddings approach outperforms them as the prediction horizon extends. The spatiotemporal positional embeddings appear to be better suited for precipitation nowcasting. Furthermore, the absence of positional encoding does not substantially degrade the model's performance, indicating that for precipitation nowcasting, positional information is only a part of the boundary conditions.

\subsection{Conclusion}

To align with the properties of meteorological tasks, this paper proposes a novel distribution-centric Meteorological Tokenization scheme for precipitation nowcasting. Depending on MeTok, the meteorological features are rearranged based on the correlations of historical information, assigning features with similar precipitation patterns to the same group. To deal with the group embeddings, the Hyper-Aligned Grouping Transformer is introduced to establish advanced aligned representations across rearranged groups while learning spatially aligned aggregations of patch embeddings within neighborhood features. Experimental evidence corroborates the efficacy of the model on precipitation nowcasting. Particularly in terms of long-tailed performance, our group learning strategy in MeTok indicates superior improvements compared to state-of-the-art methods.

\bibliography{arxiv.bib}


\appendix

\section{Appendix}

\begin{table}[h]
  \caption{The distribution of precipitation intensity on the ERA5 dataset and WeatherBench dataset. Precipitation intensity is categorized into five ascending levels: Rainless (RL), Light Rain (LR), Moderate Rain (MR), Heavy Rain (HR), and Rainstorm (RS). }
  	\renewcommand{\arraystretch}{0.7}
	\centering
    \resizebox{\linewidth}{!}
{
\begin{tabular}{c|c|c|c|c|c}
\toprule
{Intensity} & RL   & LR   & MR & HR     & RS          \\ 
Precipitation(mm) & { [}0, 0.1) & {[}0.1, 4.0) & {[}4.0, 13.0) & {[}13.0, 25.0) & {[}25.0, $\infty$) \\ 
             \midrule
ERA5         & 80.76\%      & 18.22\%        & 0.96\%          & 0.06\%           & 0.006\%              \\
WeatherBench & 83.59\%      & 15.94\%       & 0.44\%         & 0.02\%           & 0.003\%              \\ 
\bottomrule
\end{tabular}
}
  \label{tab:data}
\end{table}

\subsection{Metrics}
The threshold of precipitation is illustrated in Table~\ref{tab:data} and the derivations of metrics mentioned before are given as follows. 
The IoU is usually employed to evaluate the extent of overlap of predictions and ground truth.
In deep learning methods, IoU is an important metric to describe the extent of overlap of two regions, which is usually calculated as the following formula:
\begin{equation}
\rm IoU = \frac{predictions \cap ground truth}{predictions \cup ground truth}.
\end{equation}
Considering the inputs and outputs in our task, IoU is a very suitable evaluation metric for precipitation nowcasting.

TS is widely used in meteorology. Before giving the calculation of TS, we first introduce hits, correct negatives, false alarms, and misses:
\begin{eqnarray}
    \rm hits =  (GT>=TH)\&(pred>=TH),
\end{eqnarray}
\begin{eqnarray}
    \rm correctnegatives  =  (GT<TH)\&(pred<TH),
\end{eqnarray}
\begin{eqnarray}
    \rm falsealarms  = (GT<TH)\&(pred>=TH),
\end{eqnarray}
\begin{eqnarray}
    \rm misses  = (GT>=TH)\&(pred<TH),
\end{eqnarray}
where $\rm GT$ means groundtruth, $ \rm TH$ is threshold. After the definition of hits, correctnegatives, falsealarms, and misses, the TS, FAR, and MAR are calculated as the following formulations:
\begin{eqnarray}
    \rm TS=\frac{hits}{hits+misses+falsealarms},
\end{eqnarray}

Taking heavy rain as an illustration, we elucidate the physical significance of these metrics. The IoU is commonly utilized to assess the overlap between the predicted and actual extent of heavy rain. Threat scores (TS) are computed to compare areas where heavy precipitation is predicted with areas that experience heavy rain or more. As is the norm in related studies, we evaluate hourly predictions for each metric and report their six-hour averages in this paper.

Taking heavy rain as an illustration, we elucidate the physical significance of these metrics. The IoU is commonly utilized to assess the overlap between the predicted and actual extent of heavy rain. Threat scores (TS) are computed to compare areas where heavy precipitation is predicted with areas that experience heavy rain or more. As is the norm in related studies, we evaluate hourly predictions for each metric and report their six-hour averages in this paper.

\subsubsection{Temporal Performance Analysis}
We evaluate the model's performance in predicting precipitation at various times, as illustrated in Table~\ref{tab:overall_performances_ERA5} and Table~\ref{tab:overall_performances_WeatherBench}. Evidently, our method achieves optimal performance across all lead times in both metrics. This attests to the effectiveness of our methodology along the temporal axis as well. Particularly, there has been a substantial improvement, with a significant margin, in short-term predictions. For the precipitation at zero to one hour lead time, our approach achieves at least 7.43\% and improvement on the IoU metric over others on the ERA5 dataset. Similar performance improvement can also be observed on the WeatherBench dataset. 
Among the remaining methods, SCCN stands out as the second-performing model for predicting precipitation at zero to two-hour lead times, which also leverages historical precipitation data as prior knowledge. However, it is evident that SCCN falls short compared to our method, as SCCN's performance deteriorates rapidly with the extension of prediction time.

\begin{table}[t]\footnotesize
    \caption{Overall performances of the proposed method on the ERA5 dataset along the temporal axis.}

	\renewcommand{\arraystretch}{0.8}
	\centering
    \setlength{\tabcolsep}{1mm}
    {
		\begin{tabular}{@{}c c c c c c c c}
                \toprule
                \multirow{3}*{Metric} & \multirow{3}*{Method} & \multicolumn{6}{c}{Prediction time (hours)}\\
                \cmidrule(lr){3-8} 
                & & 0$\sim$1 & 1$\sim$2 & 2$\sim$3 & 3$\sim$4 & 4$\sim$5 & 5$\sim$6\\
                \midrule
                \multirow{13}*{IoU$\uparrow$} 
                & Pers. & 31.10 & 28.64 & 26.98 & 25.78 & 24.89 & 24.22 \\
                & Clim. & 13.36 & 13.36 & 13.36 & 13.36 & 13.36 & 13.36 \\
                & W-Clim. & 16.68 & 16.68 & 16.68 & 16.68 & 16.68 & 16.68 \\
                & ConvLSTM & 40.52 & 40.10 & 38.84 & 37.39 & 35.85 & 34.29 \\
                & ConvGRU & 40.01 & 39.71 & 38.83 & 37.75 & 36.56 & 35.31  \\
                & TrajGRU & {40.57} & {40.12} & {36.45} & {35.02} & {32.72} & 31.63 \\
                & PredRNN  & 40.56 & 39.75 & 37.49 & 35.67 & 33.59 & 32.39 \\
                & PFST & 40.67 & 40.43 & 39.29 & 37.56 & 35.94 & 34.42 \\
                & SCCN & 44.72 & 44.70 & 38.69 & 35.30 & 33.28 & 31.37\\
                & SimVP & 43.91 & 43.01 & 41.27 & 39.04 & 36.87 & 34.91\\
                & SimVP+gSTA & 44.04 & 43.01 & 40.67 & 38.25 & 36.15 & 34.65\\
                & TAU & 43.73 & 42.73 & 40.93 & 38.88 & 36.96 & 35.58\\
                & Ours & \textbf{52.15} & \textbf{48.39} & \textbf{44.83} & \textbf{42.05} & \textbf{39.49} & \textbf{37.16} 
                \\
                \cmidrule(lr){1-8} 
                \multirow{13}*{TS$\uparrow$}
                & Pers. & 15.27 & 13.57 & 12.38 & 11.125 & 10.86 & 10.34 \\
                & Clim. & 6.36 & 6.36 & 6.36 & 6.36 & 6.36 & 6.36   \\
                & W-Clim. & 7.24 & 7.24 & 7.24 & 7.24 & 7.24 & 7.24  \\
                & ConvLSTM  & 22.48 & 22.15 & 21.32 & 20.39 & 19.40 & 18.37 \\
                & ConvGRU  & 21.89 & 21.63 & 20.98 & 20.20 & 19.35 & {18.45}  \\
                & TrajGRU  & 21.87 & 21.62 & 20.22 & 18.84 & 17.61 & 16.85 \\
                & PredRNN  & 20.13 & 19.75 & 18.62 & 17.48 & 16.34 & 15.43 \\
                & PFST  & 20.06 & 20.18 & 19.58 & 18.78 & 17.95 & 17.12 \\
                & SCCN & 26.33 & 24.17 & 19.62 & 17.39 & 16.11 & 14.98 \\
                & SimVP & 22.80 & 22.53 & 21.43 & 20.01 & 18.75 & 17.65\\
                & SimVP+gSTA & 22.73 & 22.14 & 20.93 & 19.67 & 18.52 & 17.61\\
                & TAU & 22.78 & 22.17 & 21.15 & 20.11 & 19.05 & 18.20 \\
                & Ours & \textbf{28.68} & \textbf{26.06} & \textbf{23.77} & \textbf{22.08} & \textbf{20.48} & \textbf{18.96} 
                \\
                \bottomrule
		\end{tabular}}
    \centering
    \label{tab:overall_performances_ERA5}
    
\end{table}

\begin{table}[t]\footnotesize
      \caption{Overall performances of the proposed method on the WeatherBench dataset along the temporal axis.}
	\renewcommand{\arraystretch}{0.8}
	\centering
    \setlength{\tabcolsep}{1mm}
    {
		\begin{tabular}{@{}c c c c c c c c}
                \toprule
                \multirow{3}*{Metric} & \multirow{3}*{Method} & \multicolumn{6}{c}{Prediction time (hours)}\\
                \cmidrule(lr){3-8} 
                & & 0$\sim$1 & 1$\sim$2 & 2$\sim$3 & 3$\sim$4 & 4$\sim$5 & 5$\sim$6\\
                \midrule
                \multirow{13}*{IoU$\uparrow$} 
                & Pers. & 28.50 & 27.94 & 26.20 & 24.95 & 24.03 & 23.33 \\
                & Clim. & 16.10 & 16.10 & 16.10 & 16.10 & 16.10 & 16.10 \\
                & W-Clim. & 17.04 & 17.04 & 17.04 & 17.04 & 17.04 & 17.04 \\
                & ConvLSTM  & 29.56 & 29.39 & 29.05 & 28.69 & 28.25 & 27.93 \\
                & ConvGRU  & 27.64 & 27.55 & 27.23 & 26.86 & 26.53 & 26.14  \\
                & TrajGRU  & 29.32 & 29.21 & 28.85 & 28.65 & 28.24 & 27.90 \\
                & PredRNN  & 29.33 & 29.28 & 28.95 & 28.55 & 28.07 & 27.63 \\
                & PFST  & 25.96 & 25.63 & 26.08 & 26.54 & 26.87 & 26.83 \\
                & SCCN  & 30.57 & 31.27 & 31.17 & 30.76 & 30.51 & 29.57 \\
                & SimVP & 36.48 & 36.37 & 35.85 & 35.16 & 34.52 & 33.89\\
                & SimVP+gSTA & 36.23 & 36.02 & 35.44 & 34.79 & 34.07 & 33.20\\
                & TAU & 36.28 & 36.01 & 35.41 & 34.75 & 33.99 & 33.23\\
                & Ours & \textbf{41.43} & \textbf{39.40} & \textbf{37.91} & \textbf{36.75} & \textbf{35.48} & \textbf{34.41}  \\
                \cmidrule(lr){1-8} 
                \multirow{13}*{TS$\uparrow$}
                & Pers. & 14.09 & 12.18 & 10.95 & 9.86 & 9.12 & 8.54 \\
                & Clim. & 5.64 & 5.64 & 5.64 & 5.64 & 5.64 & 5.64   \\
                & W-Clim. & 5.81 & 5.81 & 5.81 & 5.81 & 5.81 & 5.81  \\
                & ConvLSTM  & 15.58 & 15.33 & 15.02 & 14.74 & 14.44 & 14.03  \\
                & ConvGRU  & 14.95 & 14.86 & 14.43 & 14.25 & 13.93 & 13.62  \\
                & TrajGRU  & 15.43 & 15.34 & 14.99 & 14.71 & {14.41} & {14.08}  \\
                & PredRNN  & 15.16 & 15.18 & 14.73 & 14.42 & 14.02 & 13.61 \\
                & PFST  & 11.14 & 10.93 & 11.39 & 11.94 & 12.27 & 12.41 \\
                & SCCN  & 15.78 & 15.87 & 15.77 & 15.73 & 15.08 & 13.82\\
                & SimVP &18.70 & 18.62 & 18.25 & 17.87 & 17.34 & 16.82 \\
                & SimVP+gSTA & 18.50 & 18.36 & 18.04 & 17.62 & 17.11 & 16.64\\
                & TAU & 18.12 & 18.05 & 17.37 & 17.12 & 17.02 & 16.20 \\
                & Ours & \textbf{21.59} & \textbf{20.54} & \textbf{19.66} & \textbf{18.89} & \textbf{18.04} & \textbf{17.30}  \\
                \bottomrule
		\end{tabular}}
        \centering
    \label{tab:overall_performances_WeatherBench}
\end{table}

\subsubsection{Comparsion with GraphCast}

\begin{table}[t]
\centering
\setlength{\tabcolsep}{0.8mm}
\caption{Comparison result of the proposed method with GraphCast on the WeatherBench 2 dataset. $\downarrow$ is better.}
{
    \begin{tabular}{lcccccc}
        \toprule
        \multirow{2.5}*{Method} & \multicolumn{6}{c}{Prediction time (h)}\\
        \cmidrule(lr){2-7}
        & 0$\sim$6& 6$\sim$12 & 12$\sim$18 & 18$\sim$24 & 24$\sim$30 & 30$\sim$36\\
        \midrule
         GraphCast& 0.57 & 0.48 &0.6 & 1.08&1.06&1.23\\
         Ours& 0.87 &1.08& 1.32& 1.47& 1.59&1.68 \\
        \bottomrule
    \end{tabular}
} 
\label{tab:graphcast}
\end{table}
We compare the proposed MeTok with SOTA data-driven method, GraphCast\cite{lam2023learning}.
Due to limited computational resources (4 Titan RTX 3090 GPUs), the parameter size of our presented model is 11.3M, while the GraphCast~\cite{lam2023learning} with 36.7M parameters is trained on 32 Cloud TPU v4 devices. Moreover, compared to our MeTok, which is trained using 79 variables with a spatial resolution of $1.5^\circ$ and a time span of 30 years, GraphCast is trained with 227 variables, including 5 surface variables and 6 atmospheric variables at each of 37 pressure levels, with a spatial resolution of $0.25^\circ$ and a time span of 39 years. Fine-grained resolution and abundant data enable Graphcast to capture the meteorological process more effectively.
However, our model still achieves performance comparable to that of GraphCast, which can be seen in Tabel~\ref{tab:graphcast}. As mentioned in the Table~\ref{tab:chap5_scal}, our model is scalable. With the increase in training data and resources, we argue that the proposed method could achieve SOTA performance.

\begin{table}[t]
\small
\caption{Comparison result of the proposed method with the baseline for 10-meter wind speed prediction on the WB2 dataset. }
\centering
\setlength{\tabcolsep}{1.mm}
{
    \begin{tabular}{lcccccc}
        \toprule
        \multirow{2.5}*{Method} & \multicolumn{6}{c}{Prediction time (h)}\\
        \cmidrule(lr){2-7}
        & 0$\sim$6 $\downarrow$& 6$\sim$12 ~$\downarrow$& 12$\sim$18 ~$\downarrow$& 18$\sim$24 ~$\downarrow$& 24$\sim$30 ~$\downarrow$& 30$\sim$36 ~$\downarrow$\\
        \midrule
         Baseline& 0.95 & 1.19 &1.45 &1.71&1.93&2.11\\
         Ours& 0.79 &0.91& 1.05& 1.18& 1.33&1.49 \\
        \bottomrule
    \end{tabular}
} 
\vspace{-0.2cm}
\label{tab:w10}
\end{table}

\noindent {\bf{Experiments on the 10-meter wind speed prediction.}} To assess the generalizability of the proposed MeTok framework, we conduct an experiment on 10-meter wind speed prediction every 6 hours within the next 36 hours. We replace the MeTok scheme with normal tokenization and the HyAGTransformer with a normal Transformer.
As shown in Table \ref{tab:w10}, the proposed method outperforms the baseline method for forecasting over all periods by a significant margin. 
The proposed method not only performs well in the precipitation prediction task but also effectively improves the performance in the near-surface wind speed prediction task. 
This further demonstrates the general effectiveness of the proposed method for meteorological forecasting tasks.


\begin{figure*}[t]
    \centering
    \includegraphics[width=\linewidth]{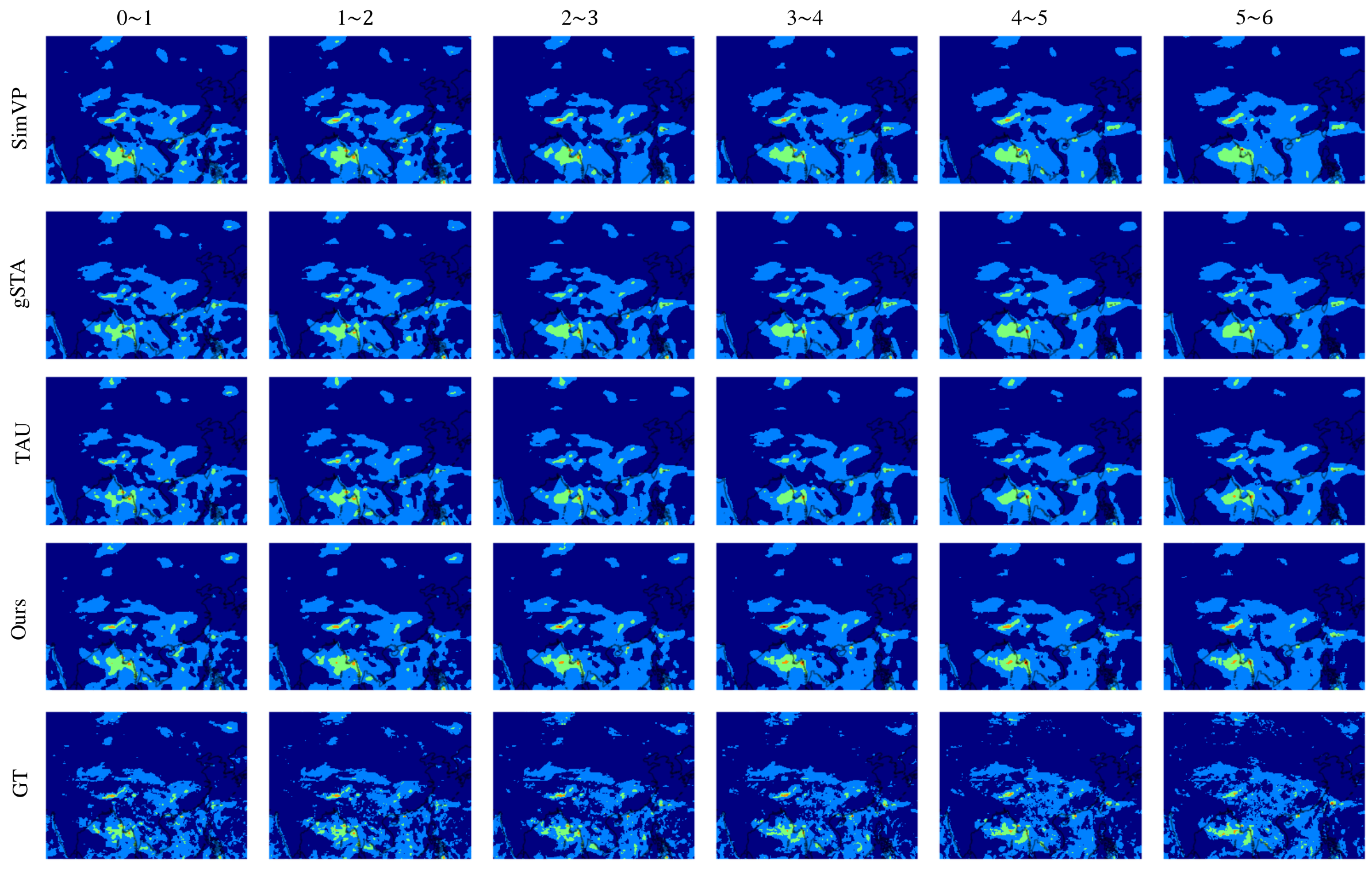}

    \caption{The visualization of prediction results on the ERA5 dataset.}
    \label{fig:result}
\end{figure*}

\subsubsection{Case Study}
This article provides a case study comparing the proposed MeTok with SimVP, SimVP+gSTA, and TAU on the ERA5 dataset, as visualized in Fig. \ref{fig:result}. 
The figure shows the precipitation forecast for the next 6 hours in East Asia on June 16, 2019, at 16:00, using the past 6 hours of meteorological data as input.
Deeper colors indicate higher precipitation levels; conversely, the lighter the color, the lower the precipitation.
For the heavy rainfall event that occurred in the Nepal region (the area on the left in the figure), only the proposed method and SimVP model are able to successfully predict it. While the gSTA and TAU barely predict this rainfall. In addition to this region, there was also a heavy rainfall event in the Myanmar region. Compared to SimVP, the proposed method successfully predicts the entire forecast period.
Overall, the proposed method has a certain improvement in the accuracy of heavy rainfall prediction, but it is still far from the actual business level. Compared to real rainfall, the rainfall predictions generated by deep learning methods are dense, but they do not perform well in predicting some discrete rainfall events.

\subsubsection{Learnable grouping strategy}
\begin{table}[t]
\centering
\caption{``MeTok" is our method, ``Learnable" indicates the learning-based grouping method. The values in bold are better.}
\setlength{\tabcolsep}{1.8mm}
\begin{tabular}{ccccccc}
     \toprule
     \multirow{3}*{Method} & \multicolumn{6}{c}{Prediction time (hours)} \\
     \cmidrule(lr){2-7}
     & 0$\sim$1 & 1$\sim$2 & 2$\sim$3 & 3$\sim$4 & 4$\sim$5 & 5$\sim$6\\
     \midrule
     
 MeTok &\textbf{52.15} & \textbf{48.39} & \textbf{44.83} & \textbf{42.05} & \textbf{39.49} & \textbf{37.16}  \\
Learnable & 40.54& 38.30& 36.93& 34.65& 33.31&32.95 \\
\bottomrule
\end{tabular}
\label{tab:2}
\end{table}

We attempted to construct an adapter to learn the ordering rules of precipitation patches. But from the results on the ERA5 dataset in Table\ref{tab:2}, it did not achieve comparable performance to our approach based on historical precipitations. However, we believe that learning-based grouping methods still hold significance in precipitation forecasting, and we plan to explore innovative and effective methods in future research. We look forward to advanced grouping designs bringing astonishing improvements in performance for precipitation nowcasting.

\begin{figure}[t]
    \centering
    \includegraphics[width=250pt]{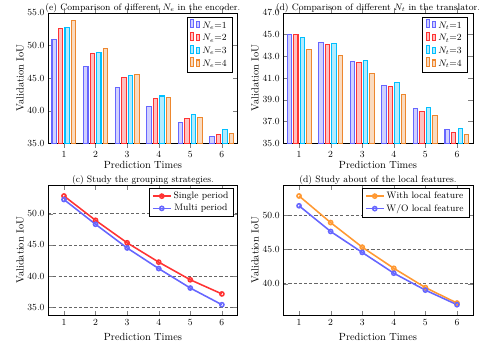}
    \caption{The ablation study of the proposed framework. We comprehensively analyzed the proposed framework, examining it from various facets.}
    \label{fig:comparison2}
\end{figure}
\begin{figure}[b!]
    \centering
    \includegraphics[width=0.7\linewidth]{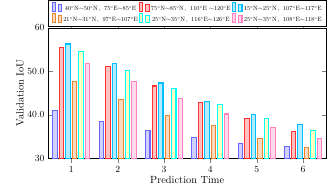}
    \caption{The predicted performance across different geographical regions.}
    \label{fig:region}
\end{figure}
\subsubsection{{Ablation study about the structure}} We analyze the effect of the number of Transformer blocks in the encoder and the translator on the model's ability to extract features. Fig. \ref{fig:comparison2} (a) and Fig. \ref{fig:comparison2} (b) show that the performance gains slowly with an increase in the number of blocks and is not directly proportional to the computational complexity. Therefore, we set $N_e$ to 4 and $N_t$ to 2 for efficiency and accuracy concurrently in this paper.

\subsubsection{{Ablation study about the grouping strategies}} In addition to utilizing recent historical precipitations as the basis for grouping, we also explore the cumulative historical precipitations over multiple periods. As shown in Fig. \ref{fig:comparison2} (c), the utilization of multi-period historical precipitations hinders the modeling of similar features.

\subsubsection{{Ablation study about of the local features}} To study the effect of the local features, experiments with or without the cross-layer connection are conducted. As shown in Fig. \ref{fig:comparison2} (d), incorporating local features with a cross-layer connection can improve the forecasting performance in the short term. 

\subsubsection{Model's adaptation across different regions}

As shown in Fig.~\ref{fig:region}, we select six regions to individually test their precipitation forecasting performance. It can be found that the performance of precipitation nowcasting in each region is relatively imbalanced, with the difference between the best and the worst results exceeding 10\%. This demonstrates that our model's adaptability needs to be strengthened, to adapt to various terrain and climate conditions. Additionally, by analyzing the precipitation results from different regions, we can see that the correlation between meteorological data and precipitation phenomena does vary by region, which is also an issue worth exploring.


\end{document}